\documentclass[pdflatex,sn-mathphys]{sn-jnl}

\usepackage{array}
\usepackage{tabularx}
\usepackage{multirow}
\usepackage{booktabs}
\jyear{2021}%

\theoremstyle{thmstyleone}%
\theoremstyle{thmstyletwo}%

\theoremstyle{thmstylethree}%

\begin{document}

\title[Article Title]{A New Spatial Adjacency Matrix of Skeleton Data Based on Self-loop and Adaptive Weights}


\author[1]{\fnm{Zheng} \sur{Fang}}

\author[1]{\fnm{Xiongwei} \sur{Zhang}}

\author*[1]{\fnm{Tieyong} \sur{Cao}}\email{cty\_ice@sina.com}

\author[1,2]{\fnm{Yunfei} \sur{Zheng}}

\author[1]{\fnm{Meng} \sur{Sun}}

\affil*[1]{\orgdiv{Institute of Command and Control Engineering}, \orgname{Army Engineering University}, \orgaddress{\city{Nanjing}, \postcode{210001}, \state{Jiangsu}, \country{China}}}

\affil[2]{\orgdiv{The Army Artillery and Defense Academy of PLA}, \orgaddress{\city{Nanjing}, \postcode{211131}, \state{Jiangsu}, \country{China}}}


\abstract{Human skeleton data has received increasing attention in action recognition due to its background robustness and high efficiency. In skeleton-based action recognition, graph convolutional network (GCN) has become the mainstream method. This paper analyzes the fundamental factor for GCN-based models -- the adjacency matrix. We notice that most GCN-based methods conduct their adjacency matrix based on the human natural skeleton structure. Based on our former work and analysis, we propose that the human natural skeleton structure adjacency matrix is not proper for skeleton-based action recognition. We propose a new adjacency matrix that abandons all rigid neighbor connections but lets the model adaptively learn the relationships of joints. We conduct extensive experiments and analysis with a validation model on two skeleton-based action recognition datasets (NTURGBD60 and FineGYM). Comprehensive experimental results and analysis reveals that 1) the most widely used human natural skeleton structure adjacency matrix is unsuitable in skeleton-based action recognition; 2) The proposed adjacency matrix is superior in model performance, noise robustness and transferability.}

\keywords{Skeleton-Based action recognition, Graph convolutional network, Adjacency matrix, Human natural skeleton structure, Noise robustness}



\maketitle

\section{Introduction}\label{sec1}

Action recognition is a fundamental task in video understanding with a wide range of practical applications, including security, human-computer interaction, robotics\cite{23}\cite{24}\cite{25}, etc. Skeleton-based action recognition has drawn much attention recently. Compared with RGB representations, skeleton data is more robust to appearances and requires fewer computational resources. Researchers have proposed many deep models based on convolutional neural network (CNN)\cite{7}, recurrent neural network (RNN)\cite{4}, and GCN \cite{10} for skeleton-based action recognition. Generally speaking, the GCN-based model is the best model for its superior performance and efficiency.

The typical GCN-based models\cite{10}\cite{11}\cite{13}\cite{40} treat skeleton data as graph structures, the body joints are graph nodes and connections between joints are graph edges. The graph sequences are processed by graph convolution to predict action categories. The adjacency matrix, which represents the edges between nodes, is crucial for graph data. ST-GCN\cite{10} first uses the natural connections of human body joints to form the adjacency matrix $A_{Skeleton}$ (Fig. 1) to represent the spatial edges of joints. Following ST-GCN, lots of different GCN-based models\cite{11}\cite{13}\cite{40} have also been proposed. Although these models propose many different adjacency matrices, $A_{Skeleton}$ is still the first and necessary choice in these methods.
\begin{figure}[htb]
\centering
\includegraphics[width=3in]{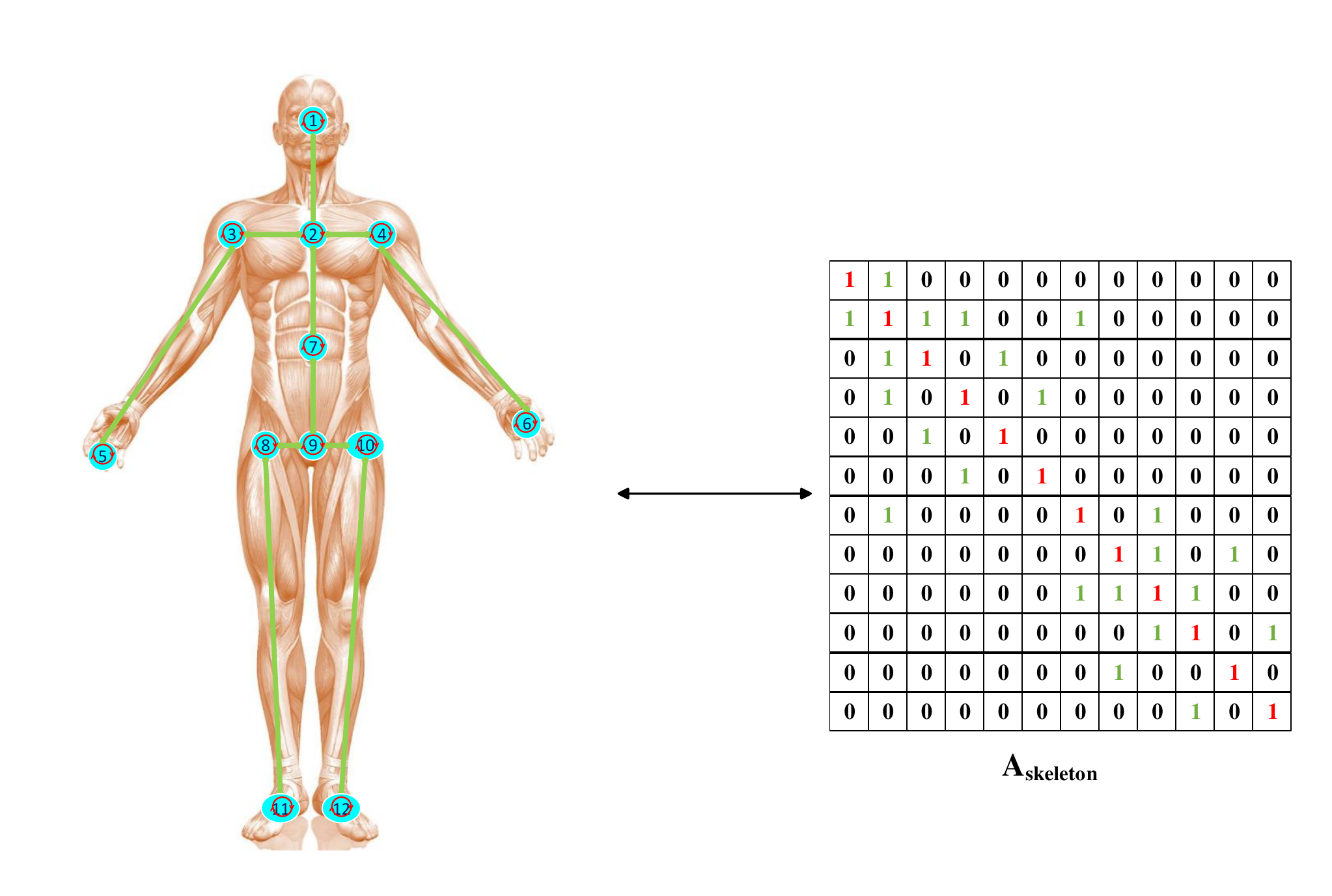}
\caption{A 12-joint skeleton structure and its corresponding adjacency matrix $A_{Skeleton}$}
\label{Fig1}
\end{figure}

The $A_{Skeleton}$ seems to be a reasonable choice for the natural structure of body joints, and no researchers have thought and questioned it before. However, we find that the final adjacency matrices used in recent works\cite{11}\cite{17} are very far from the original $A_{Skeleton}$. Besides, in our former work\cite{41}, we just use the identity matrix $I$ to replace $A_{Skeleton}$ and the model obtains better performance. These results attract our attention, and we argue that is $A_{Skeleton}$ necessary in GCN models for skeleton-based action recognition?

In this paper, we deeply analyze the role of $A_{Skeleton}$ in the GCN model. We propose that $A_{Skeleton}$ is not a proper choice in GCN models for skeleton-based recognition due to the existence of $A_{Sk-neighbor}$ (the green edges in Fig. 1). Besides, based on the results of our former work \cite{41}, we propose a new adjacency matrix to replace $A_{Skeleton}$. We set $A=I+A_{res}$ where $I$ is the identity matrix and $A_{res}$ is an adaptively learned matrix by the model. The proposed adjacency matrix focuses on the motion feature of the joint itself, removes all rigid connections between different joints but lets the model adaptively learn relationships of joints.

Specifically speaking, two works are mainly carried out in this study:

(1) Five most typical models are firstly selected to verify our idea preliminarily. Then, a simple GCN model is built, and extensive experiments are conducted on two kinds of skeleton-based action recognition datasets, to deeply study the influence of the adjacency matrix. Experimental results strongly support our idea that $A_{Skeleton}$ is not a proper choice, and the proposed adjacency matrix is better than $A_{Skeleton}$.

(2) We analyze the advantages of the proposed adjacency matrix compared to $A_{Skeleton}$. By comparing the performance of different adjacency matrices under different noise, we find that the proposed adjacency matrix is more robust to noisy skeleton data. Besides, the proposed adjacency matrix has better transferability for different types of skeleton data.

The important contribution of this paper is that our experimental results and analysis fully suggest that $A_{Skeleton}$ is an unsuitable choice, and we provide a relatively better adjacency matrix. For skeleton-based action recognition, our conclusions are beneficial for future GCN-based models and can also be considered by other types of deep models.

\section{Related Work}\label{sec2}
\textbf{Skeleton-based action recognition models}: In CNN-based models\cite{7}\cite{8}, skeleton data is usually transformed into image sequences to be processed by CNNs. RNN-based models treat skeleton data as temporal sequences\cite{4} to explore the temporal dynamics of joints. Recently, GCN-based models\cite{10}\cite{11}\cite{13}\cite{40}\cite{54}\cite{55} have become the dominant approaches to process skeleton data due to the superior performance and efficiency. Yan et al.\cite{10} first propose ST-GCN, skeleton data are represented as graph data, and natural skeleton connections are used to build adjacency matrix of each skeleton graph. Shi et al.\cite{11} propose 2s-AGCN which fuses the joint-stream and bone-stream to boost performance. Cheng et al.\cite{15} propose Shift-GCN which is computationally efficient and provides flexible receptive fields for both spatial graph and temporal graph. GR-GCN\cite{16} and MS-G3D\cite{17} build spatial-temporal skeleton graphs to directly model cross-spacetime joint dependencies. Wu et al.\cite{50} propose Graph2Net which uses two different GCNs to process spatial and temporal graphs.

\textbf{The adjacency matrix in GCN models}:
ST-GCN\cite{10} first uses $A_{Skeleton}$ to represent the spatial edges of joints. Following ST-GCN, most subsequent works \cite{15}\cite{14}\cite{16}\cite{49}\cite{50}\cite{40} also use $A_{Skeleton}$ as their main adjacency matrix. Some recent works perform some operations on $A_{Skeleton}$. For example, 2s-AGCN \cite{11} adds self-attention coefficients and a freely learned adjacency matrix on $A_{Skeleton}$ to improve performance. In \cite{26}, the final adjacency matrices are composed of the $1_{th} ... K_{th}$ powers of  $A_{Skeleton}$. MS-G3D\cite{17} uses $1_{th} ... K_{th}$ disentangled aggregations of $A_{Skeleton}$ to perform multi-scale graph convolution.

In general, a series of recent works have advocated many innovations in designing adjacency matrix, however, no matter how the adjacency changes in these GCN models, $A_{Skeleton}$ is always the first choice. In our former work STSF-GCN\cite{41}, a slowfast spatial-temporal GCN is proposed and the model perform better when replacing $A_{Skeleton}$ by identity matrix $I$. In this paper, based on some conclusions and data of STSF-GCN, we conduct a more in-depth and comprehensive analysis of the adjacency matrix. Unlike most former works, a new adjacency matrix which does not contain $A_{Skeleton}$ is proposed in our work. A large number of experimental results and analysis verify that the proposed adjacency matrix is better than $A_{Skeleton}$ in model performance, noise robustness and data transferability.

\section{Preliminary experiments on typical GCN models}\label{sec3}
The idea of this paper comes from our former work \cite{41}. In \cite{41} we propose STSF-GCN for skeleton-based action recognition, and we find that the STSF-GCN performs better when replacing $A_{Skeleton}$ by identity matrix $I$. The result raises our interest in the effectiveness of $A_{Skeleton}$ in GCN-models. In order to eliminate the influence of model structures or data preprocessing methods, we conduct further experiments on four most recent and typical GCN models: ST-GCN\cite{10}, JS-AGCN\cite{11}, MS-G3D\cite{17}, Shift-GCN\cite{15}. All these models use $A_{Skeleton}$ as the main adjacency matrix. We use the code released by authors and just replace $A_{Skeleton}$ by $I$ to train these models on NTU-RGBD dataset\cite{28} (The most commonly used skeleton-based action recognition dataset). The results are shown in Table 1 and we also add the results of our former model STSF-GCN.

\begin{table}[h]
\newcommand{\tabincell}[2]{\begin{tabular}{@{}#1@{}}#2\end{tabular}}
\caption{\upshape Accuracy(\%) of some typical GCN models using different adjacency matrix, CS and CV are different setting in NTU-RGBD dataset}
\centering
\begin{tabular}{cccc}
\toprule
Model&Adjacency matrix&\tabincell{c}{NTU-\\RGBD-CS}&\tabincell{c}{NTU-\\RGBD-CV} \\
\midrule
\multirow{2}{*}{STGCN}&$\underline{A=A_{Skeleton}}$&78.72&86.3  \\
&$A=I$&79.68&87.1\\
\midrule
\multirow{2}{*}{JS-AGCN}&$\underline{A=A_{Skeleton}}$&85.7&93.7  \\
&$A=I$&85.6&93.5\\
\midrule
\multirow{2}{*}{MS-G3D}&$\underline{A=A_{Skeleton}}$&88.8&95.0  \\
&$A=I$&88.9&95.0\\
\midrule
\multirow{2}{*}{Shift-GCN}&$\underline{A=A_{Skeleton}}$&85.4&91.5  \\
&$A=I$&85.5&91.6\\
\midrule
STSF-GCN&$\underline{A=A_{Skeleton}}$&89.0&95.0  \\
(our former work)&$A=I$&89.5&95.4\\
\bottomrule
\end{tabular}
\end{table}

As Table 1 shows, in these GCN models, setting $A=I$ or $A_{Skeleton}$ does not have much difference, $A=I$ even better than $A_{Skeleton}$ in some cases. Combining the example in Fig. 1, we can preliminarily think that the $A_{Sk-neighbor}$ is not helpful for skeleton-based recognition.

\section{Task analysis}\label{sec4}
\subsection{Preliminaries}\label{subsec2}
In one frame, a human skeleton graph is denoted as $G=(V, \varepsilon)$, $V$ is the set of $N$ joints, and $\varepsilon$ is the edge set. $X \in R^{T \times N \times C}$ represents $T$ skeleton graphs, where $T$ is the number of frames and $C$ is the feature dimension of each joint. For a single skeleton graph $X_t \in R^{N \times C}(0 < t \le T)$ and its corresponding adjacency matrix $A \in R^{N \times N}$, the graph convolution operation is computed by:
\begin{equation}
X_t^{L+1}=\sigma(\bar{A}X_t^LW^L),
\end{equation}
where $\bar{A}$ is the normalization matrix of $A$, $W^L$ is the $L_{th}$ layer's weight of the $1 \times 1$ convolution, $\sigma$ is an activation function. Most GCN-based methods\cite{10}\cite{11}\cite{14} model skeleton data in one frame as a spatial graph. In these methods, Eq. (1) is spatial graph convolution and $A$ represents the spatial relationships of joints. Some methods\cite{16} \cite{17} model skeleton data in a unified spatial-temporal way. In this case, $A$ becomes a block matrix whose diagonal sub-matrices represent the spatial relationships ($A_{Spatial}$) of the joints in the same frame, and the non-diagonal sub-matrices represent the temporal relationships ($A_{Temporal}$) between the joints in different frames.
\subsection{Analysis on $A_{Skeleton}$}
In Eq. (1), despite the activation function $\sigma$ and layer's weight $W^L$, the key factor lies in $\bar{A}X_t^L$ for graph data. For the convenience of discussion, assuming $N=3$, for one graph convolution layer, the input data is
\begin{equation}
X^L_t=\left(
  \begin{array}{ccc}
    x^{L}_{t,1}\\
    x^{L}_{t,2}\\
    x^{L}_{t,3} \\
  \end{array}
\right).
\end{equation}
The normalized adjacency matrix is:
\begin{equation}
\bar{A}=\left(
  \begin{array}{ccc}
    a_{11}&a_{12}&a_{13} \\
    a_{21}&a_{22}&a_{23} \\
    a_{31}&a_{32}&a_{33} \\
  \end{array}
\right),
\end{equation}
so
\begin{equation}
\bar{A}X^L_t=\left(
  \begin{array}{ccc}
    a_{11}x^{L}_{t,1}+a_{12}x^{L}_{t,2}+a_{13}x^{L}_{t,3} \\
    a_{21}x^{L}_{t,1}+a_{22}x^{L}_{t,2}+a_{23}x^{L}_{t,3} \\
    a_{31}x^{L}_{t,1}+a_{32}x^{L}_{t,2}+a_{33}x^{L}_{t,3} \\
  \end{array}
\right).
\end{equation}

Comparing Eq. (4) and Eq. (2), for the first node, its feature has changed from $x^{L}_{t,1}$ to $a_{11}x^{L}_{t,1}+a_{12}x^{L}_{t,2}+a_{13}x^{L}_{t,3}$. The discussion of the adjacency matrix is actually to analyze the rationality of the coefficients $a_{11}, a_{12}, a_{13}$. That is, can the features weighted by these coefficients help the network identify the action category more accurately?

On the other hand, skeleton-based action recognition is very different from most typical GCN tasks \cite{18}. As the bottom part of Fig. 2 shows, the previous mainstream works on GCN mainly focused on every single node and its adjacent nodes. The entire graph structure is fixed during processing, i.e., a map from a single graph to another single graph. But for the skeleton-based action recognition, as the top of Fig. 2 shows, the action category is classified by the aggregation of all features of all joints from all frames in a period of action. No matter what kind of GCN models are used, skeleton-based action recognition needs to aggregate these features into one vector, i.e., a map from multiple graphs to one classification vector.

\begin{figure}[htb]
\centering
\includegraphics[width=3.5in]{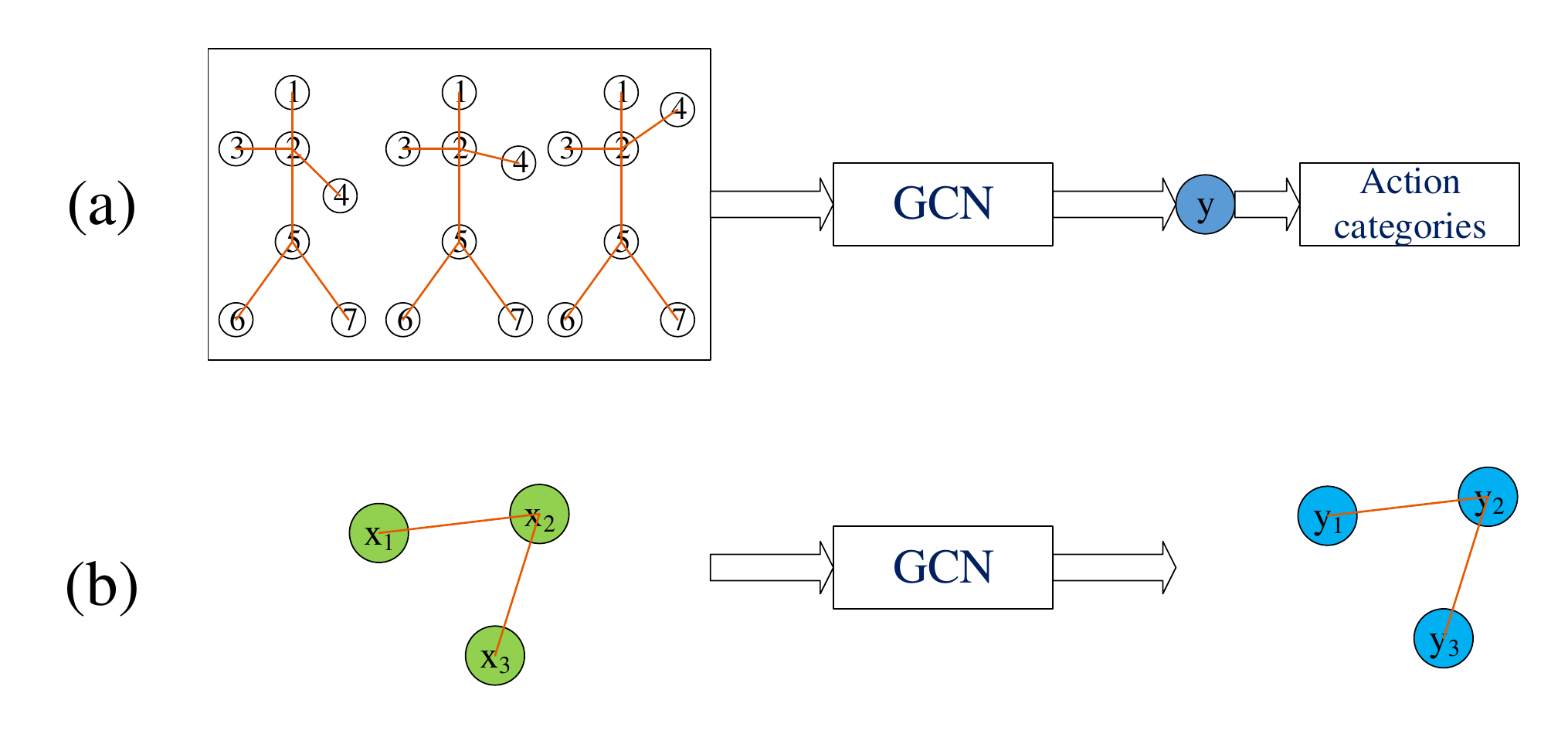}
\caption{Visualization of skeleton-based action recognition (a) and the typical GCN task (b)}
\label{Fig2}
\end{figure}

Generally speaking, most models perform average or max pooling operations on both space and time dimensions to fuse the features of joints. Suppose that the spatial-temporal average pooling results of equation (4) are used to classify action category, the final output vector is
\begin{equation}
F_{Cls}= \frac{\sum_{t=1}^T \sum_{n=1}^N(\bar{A}X_t^L)}{NT} \in R^{1 \times C_{Action}},
\end{equation}
where $n$ is the spatial identifier of each joint and $C_{Action}$ is the number of action categories. From Eq. (5), we can find that the model needs an adjacency matrix that properly represents relationships of all joints' motion features to help the model identify the action category more accurately.

Based on the above analysis, we consider that $A_{Skeleton}$ is not a proper choice in GCN models for skeleton-based recognition, and the problem is $A_{Sk-neighbor}$ which represents the neighbor connections between human joints. There are two reasons:

(1) $A_{Sk-neighbor}$ is an ideal symmetric non-negative adjacency matrix, so the coefficients $a_{11}, a_{12}, a_{13}$ in Eq. (4) is always fixed and non-negative. However, due to the diversity of action categories, the relationships between joints are different for different actions. \textbf{Using $A_{Sk-neighbor}$ can not guarantee a positive effect to classify all actions for GCN models.}

(2) In skeleton-based action recognition, the model uses the adjacency matrix to fuse all joints' motion features. But as Fig. 1 shows, $A_{Sk-neighbor}$ \textbf{only represents the relative positional relationships of different joints in the same frame. These relative positional relationships may not accurately describe the motion information association between different joints during the action.}

Combining all the above analysis and results from our former work\cite{41}, we propose using $A=I+A_{res}$  to replace $A_{Skeleton}$. $I$ is the identity matrix, and $A_{res}$ is a $N \times N$ adaptively learned matrix, and the elements of $A_{res}$ are parameterized and optimized together with the other parameters in the training process. The proposed adjacency matrix focuses on the motion feature of the joint itself. It removes all rigid connections between different joints but lets the model adaptively learn the relationships of joints. The adaptively learned matrix is also used in other works\cite{11}\cite{17}, the core innovation of our work is actually the factor $I$ that brings many advantages. In the rest of this paper, extensive experiments are designed to demonstrate our conclusions and the effectiveness of the proposed adjacency matrix.
\section{Settings on validation model and datasets}

In order to deeply analyze the adjacency matrix in GCN models. We conduct a simple GCN validation model to perform comparative experiments and analysis. The model is shown in Fig. 3, and its structure is similar to the fast pathway in STSF-GCN\cite{41}.
\begin{figure}[htb]
\centering
\includegraphics[width=3.5in]{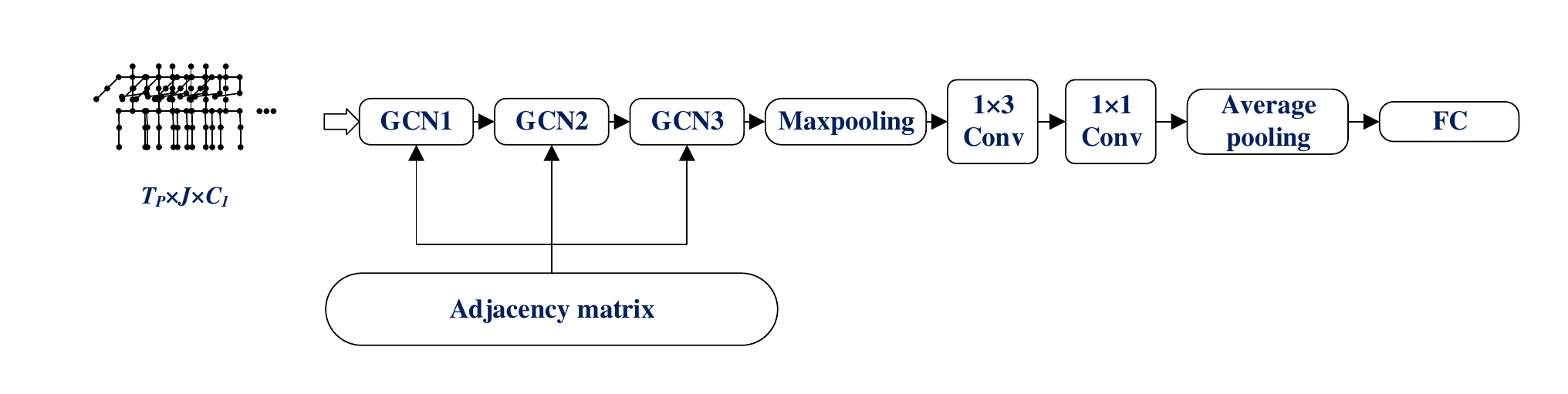}
\caption{The model used to perform comparative experiments and analysis, GCNs are layers of graph convolution and FC denotes the fully-connected layer.}
\label{Fig2}
\end{figure}
\begin{figure}[htb]
\centering
\includegraphics[width=4in]{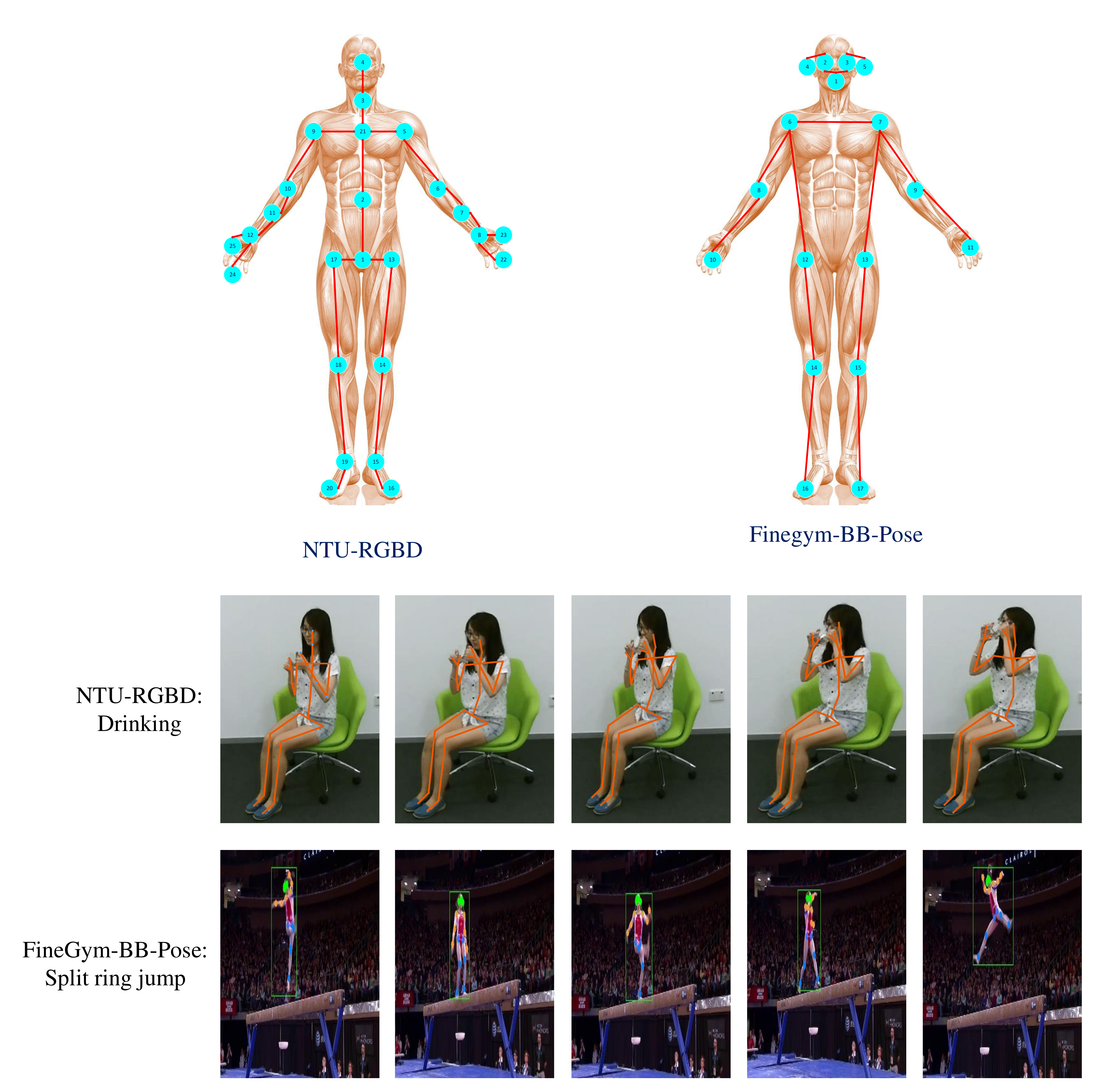}
\caption{Visualizations of joints and action examples in NTU-RGBD and FineGym-BB-Pose}
\label{Fig2}
\end{figure}
The input of the model are sequences of skeleton data, and each joint contains its coordinate feature and velocity feature. It contains 3 graph convolution layers, several convolution layers, pooling layers and a fully-connected layer to process skeleton data. We choose this validation model because it is a typical GCN structure. It is simple and high-efficiency, which is very helpful for us to perform experiments and analysis. Details of the model can refer to \cite{41}.

For skeleton-based action recognition datasets, considering different types of skeleton data and actions, we choose the following two datasets to train and test model.

(1) \textbf{NTU-RGBD}. NTU-RGBD\cite{28} is the most widely used dataset in skeleton-based action. It contains 56,880 skeleton action sequences performed by 40 volunteers with 3 camera views and categorized into 60 classes. Each skeleton sample contains 25 joints captured by Microsoft Kinect v2 cameras. The actions in the NTU-RGBD are generally daily actions like eating or drinking. We use the CS setting of NTU-RGBD where the 40 subjects are split into training and testing groups, yielding 40,091 and 16,487 training and testing examples, respectively.

(2) \textbf{FineGym-BB-Pose}. FineGym-BB-Pose is a skeleton dataset conducted by us. It is obtained by performing pose estimation models on FineGym-BB\cite{42} RGB video dataset. FineGym-BB is a video action recognition dataset containing many professional balance beam actions. We use the pretrained Mask-RCNN\cite{43} and HR-Net\cite{44} on original RGB frames of FineGym-BB to get the 2D skeleton data. Each skeleton sample contains 17 joints, and the dataset contains 7037 train samples and 3035 test samples. We choose this dataset because the actions in FineGym-BB can be classified only by the skeleton features of athletes regardless of the backgrounds or other objects. In this dataset, the 2D skeleton data is easier to estimate compared with other RGB action recognition datasets, which ensures the quality of the skeleton data.

The skeleton nodes and action examples of the above two datasets are shown in Fig. 4.

\textbf{Implementation Details:} The model is trained on Pytorch platform with one RTX2080TI. We use the Adam [20] optimizer with the initial learning rate of 0.001. The learning
rate decays by a factor of 10 at $40^{th}$ epoch and $80^{th}$ epoch. The training is finished at $120^{th}$ epoch. The weight decay is 0.0001. The batch sizes are set to 32 for all datasets. We adopt the data pre-processing method in \cite{13}. Label smoothing is utilized for all experiments and the smoothing factor is 0.05. Cross-entropy loss is used to train the networks. The input frames are $160$ and $60$ for NTU-RGBD and FineGym-BB-Pose, respectively.

\section{Experiment results and analysis}
\subsection{The influence of adjacency matrix on model performance}
We focus on the influence of the adjacency matrix in GCN1-3 in Fig. 3. We first use the model train and test on two datasets with different adjacency matrix settings. Both spatial graph convolution and spatial-temporal graph convolution are tested (the specific details of spatial-temporal graph convolution can refer to \cite{41}). Results are shown in Table 2 (spatial graph convolution) and Table 3 (spatial-temporal graph convolution).
\begin{table}[h]
\newcommand{\tabincell}[2]{\begin{tabular}{@{}#1@{}}#2\end{tabular}}
\caption{\upshape Accuracy(\%) of the validation model using different adjacency matrices, GCN1-3 are spatial graph convolution layers, $A_{res}$ is an adaptively learned matrix}
\centering
\begin{tabular}{cccc}
\toprule
No.&Adjacency matrix&NTU-RGBD-CS&\tabincell{c}{FineGym\\-BB-Pose} \\
\midrule
1&$A=A_{Skeleton}$&83.5&80.4  \\
2&$A=I$&83.5&80.5\\
3&$A=A_{Sk-neighbor}$&82.5&79.9\\
4&$A=I+A_{res}$&\textbf{85.6}&\textbf{82.2}\\
5&$A=A_{Skeleton}+A_{res}$&84.7&81.4\\
\bottomrule
\end{tabular}
\end{table}

\begin{table}[htb]
\newcommand{\tabincell}[2]{\begin{tabular}{@{}#1@{}}#2\end{tabular}}
\caption{\upshape Accuracy(\%) of the validation model using different adjacency matrices, GCN1-3 are spatial-temporal graph convolution layers, $\tau$ is the number of frames for building spatial-temporal graph}
\centering
\begin{tabular}{ccccc}
\toprule
No.&$\tau$&Adjacency matrix&NTU-RGBD-CS&\tabincell{c}{FineGym\\-BB-Pose} \\
\midrule
1&3&\tabincell{c}{$A_{Spatial}=A_{Skeleton}$,\\ \underline{$A_{Temporal}=I$}}&85.1&79.6  \\
2&3&\tabincell{c}{$A_{Spatial}=I$,\\ \underline{$A_{Temporal}=I$}}&85.2&79.8\\
3&5&\tabincell{c}{$A_{Spatial}=A_{Skeleton}$,\\ \underline{$A_{Temporal}=I$}}&85.0&79.9\\
4&5&\tabincell{c}{$A_{Spatial}=I$,\\ \underline{$A_{Temporal}=I$}}&85.6&80.4\\
\bottomrule
\end{tabular}
\end{table}

Firstly, the results of Table 2 and Table 3 show that $A=I$ is better than $A=A_{Skeleton}$ in both spatial graph convolution and spatial-temporal graph convolution, which is consistent with the results in Table 1. Since the temporal relationships in spatial-temporal graph convolution are not the main content of our work, we just further analyze the results in Table 2 to study the influence of the adjacency matrix. Two conclusions can be obtained from Table 2:

(1) In Table 2, compared with model 1 and 2, model 3 gets the lowest accuracy, which demonstrates that $I$ plays a major role in recognition.

(2) Setting $A=I+A_{res}$ can let the model get the best performance (model 4 in Table 2).

Two key questions are also derived from the above two conclusions: (1) What is the influence of $A_{Sk-neighbor}$ in action recognition? (2) What kind of adjacency matrix is suitable for skeleton-based action recognition? That is, what does the learned $A_{res}$ in model 4 look like? We conduct a more in-depth analysis next to respond to these two questions.
\subsection{Influence of $A_{Sk-neighbor}$}
Previous research \cite{20} shows in theory that the graph convolution is actually a Laplacian smoothing filter\cite{22} applied on the feature matrix for low-pass denoising. Combining Eq. (1) and Eq. (4), if we use $A_{Sk-neighbor}$ to perform graph convolution, the feature of the joint itself will be smoothed by its neighbor joints because all coefficients are non-negative. But in skeleton-based action recognition, some actions are only performed by a few joints. For example, ``Reading", ``clapping" and ``playing cell-phone" require only hand joints to complete. With $A_{Sk-neighbor}$, the moving joints are connected with the static joints. The high-frequency motion feature may be restrained in graph convolution, which will mislead the model.

We choose model 2 and model 3 in Table 2 to prove the above analysis. We count their classification results in NTU-RGBD and find that 836 samples are correctly classified by model 2 but are misclassified by model 3. Fig. 5 shows the Top-15 action categories of these 836 samples.
\begin{figure}[htb]
\centering
\includegraphics[width=3in]{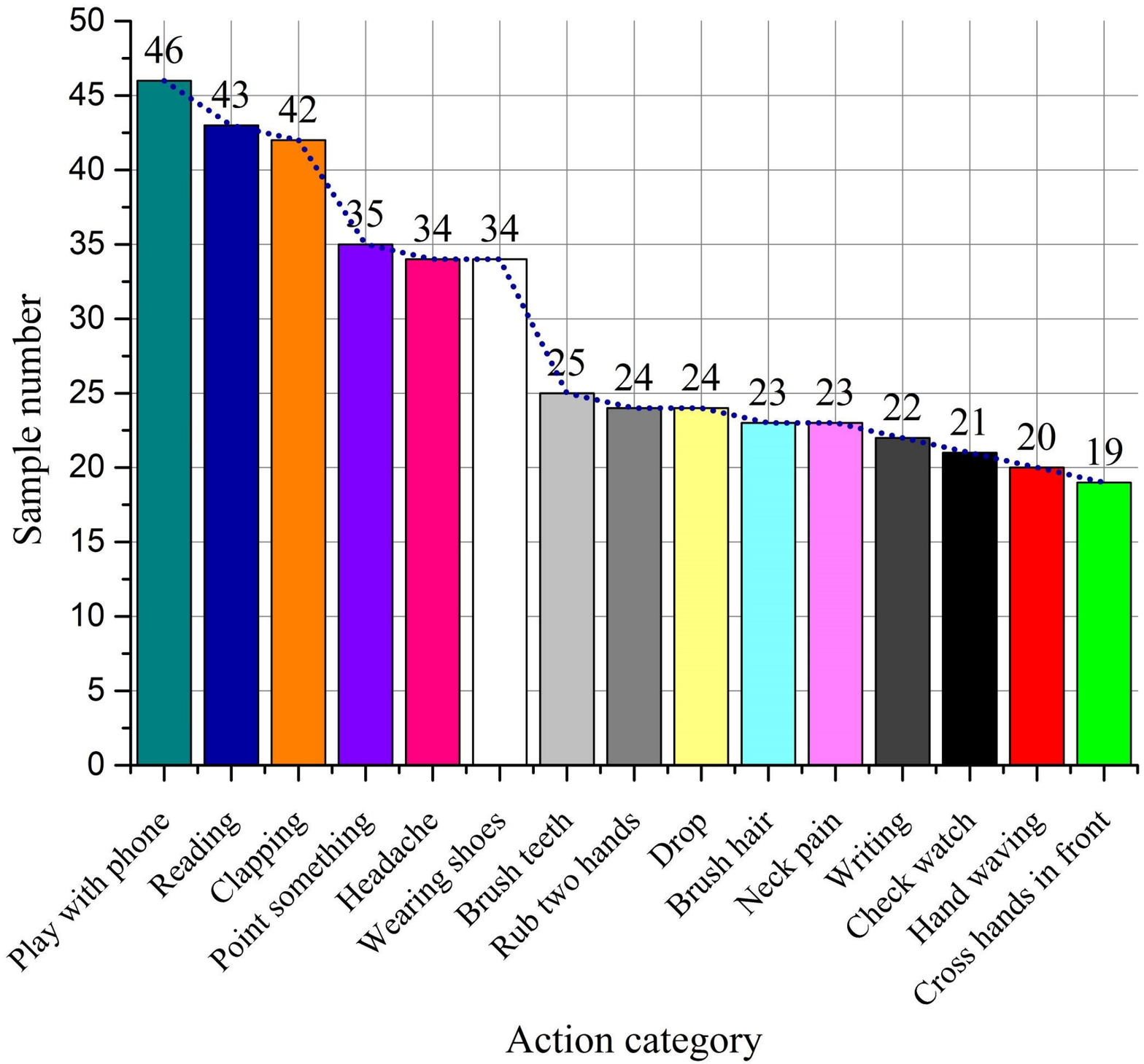}
\caption{Top-15 action categories that are correctly classified by the model 2 (Table 2) but are misclassified by the model 3 (Table 2).}
\label{Fig2}
\end{figure}

As Fig. 5 shows,  actions ``Play with phone", ``Reading" , ``Clapping", ``Point something", ``Headache" are most likely to be misclassified by the model 3 in Table 2. These actions are all performed by a few joints and the motion of each joint is tiny. The number of them accounted for 24\% of the total, since NTU-RGBD contains 60 action classes, the rate of 24\% can clearly show the weakness of model 3.
\begin{figure*}[htb]
\centering
\includegraphics[width=4in]{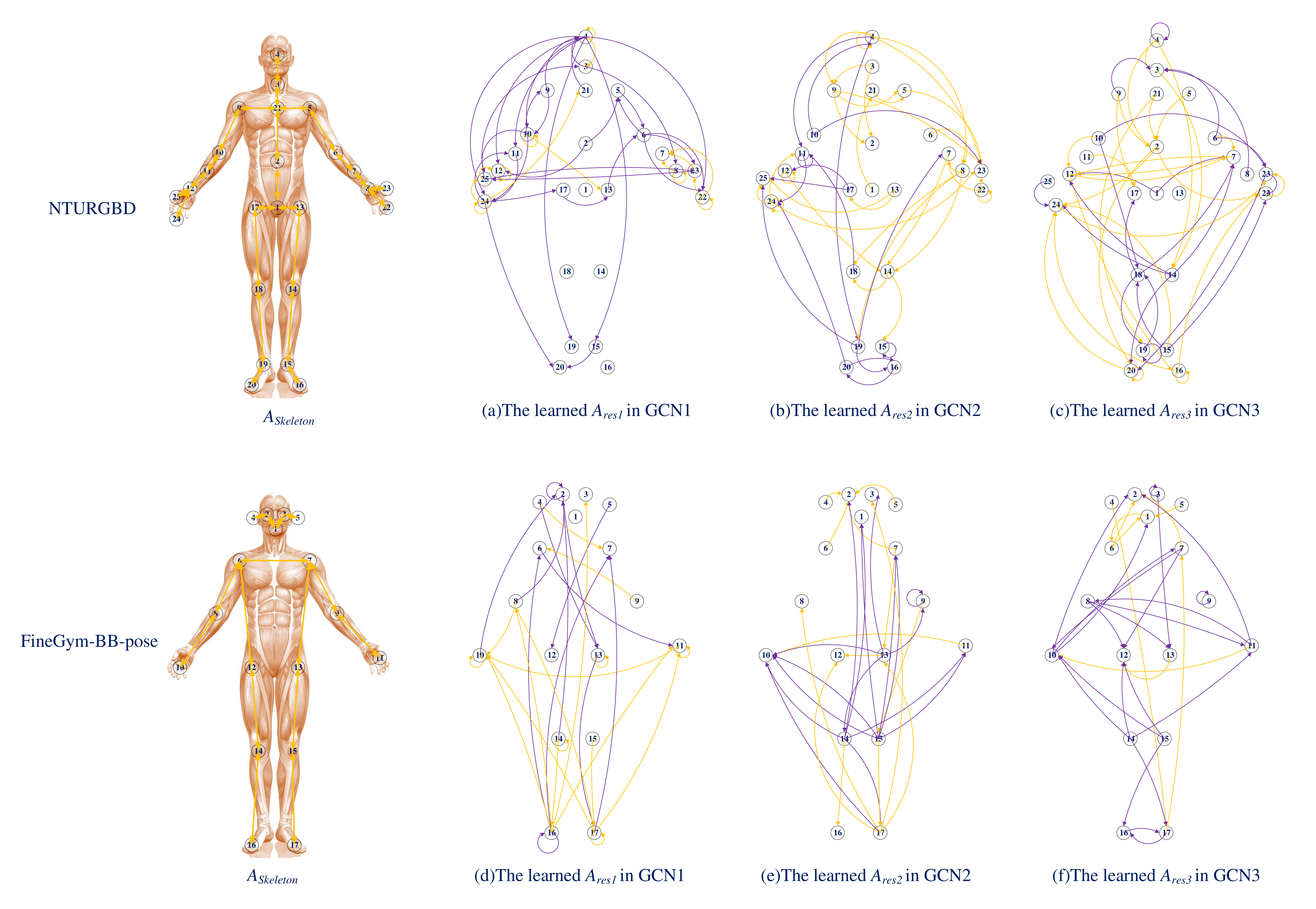}
\caption{The visualization of $A_{Skeleton}$  and the learned adjacency matrices $A_{res1}, A_{res2}, A_{res3}$ of model 4 in Table 2. Yellow lines represent positive edges and purple lines represent negative edges.}
\label{Fig6}
\end{figure*}
\subsection{Analysis on $A_{res}$}
In Table 2, the accuracy of model 4 and model 5 is much higher than other models, especially model 4. Here we simply think that the $A_{res}$ learned in model 4 is a ``correct" adjacency matrix. In order to study $A_{res}$ more intuitively, we plot the learned $A_{res}$ in GCN1-3 of model 4 in Fig. 6. We refer to $A_{Skeleton}$ when plotting $A_{res}$, for dataset NTU-RGBD, the $A_{Skeleton}$ contains 48 edges totally, so we choose the first 48 edges ordered by the absolute values in $A_{res}$; For dataset FineGym-BB-Pose, the $A_{Skeleton}$ contains 30 edges totally, so we choose the first 30 edges ordered by the absolute values in $A_{res}$.

We can get much valuable information from Fig. 6:

(1) $A_{Skeleton}$ is a symmetric matrix, but the learned $A_{res}$ is not symmetric. This fact indicates that the relationships of joints are not always bidirectional.

(2) In $A_{Skeleton}$, the weights of all edges are positive, but there are many negative edges in $A_{res}$. It also demonstrates that the edges in $A_{Skeleton}$ are unsuitable.

(3) The learned $A_{res1}, A_{res2}, A_{res3}$ in GCN1-3 are different. It indicates that the ``correct" relationships between joints will change with the network going deeper. This is a problem that has been ignored in most previous work\cite{10}\cite{11}\cite{15}\cite{17}\cite{40}, in these works, all graph convolution layers use $A_{Skeleton}$ as their necessary choice.

(4) Although the adjacency matrix $A=I+A_{res}$ contains the fixed part $I$, there are still many self-loop edges in $A_{res}$. For example, No. 10, 12, 24, 8, 22, 4 joints in part (a) and No. 10, 13, 11, 14, 17 joints in part (d). This phenomenon also proves the effectiveness of our proposed matrix.

From the perspective of the dataset, we can also get much information. For example, in the NTU-RGBD dataset, many actions are performed by the joints of the head and hand, such as eating and brushing teeth. So there are many connections between head and hand joints. For the FineGym-BB-Pose dataset, actions are completed on the balance beam. Most of them are completed by the force of the foot, so the edges are mostly related to the foot joints. Lots of information can also be obtained if we concentrate on specific actions.

Based on the above experiment results and analysis, we can get the conclusion that $A_{Skeleton}$ is not a proper adjacency matrix in skeleton-based action recognition. Due to the wide range of action types, fixed weights of edges are not suitable. In Table 2, model 5 ($A=A_{Skeleton}+A_{res}$) performs worse than model 4 ($A=I+A_{res}$) even with same factor $A_{res}$. The rigid connections in $A_{Skeleton}$ prevent the GCN model from learning the optimal edges. Setting $A=I+A_{res}$ is a better choice.
\section{Noise robustness and transferability of the proposed adjacency matrix}
Results and analysis in Sec. 6 show that the proposed adjacency matrix is better than $A_{Skeleton}$ in terms of the model performance. In fact, the proposed adjacency matrix also has advantages in noise robustness and transferability. The main reason is that the fixed part $I$ is less sensitive to some noise and different types of skeleton data. In this section, we conduct experiments and analysis to show the superiority of the proposed adjacency matrix.
\begin{figure*}[htb]
\centering
\includegraphics[width=4in]{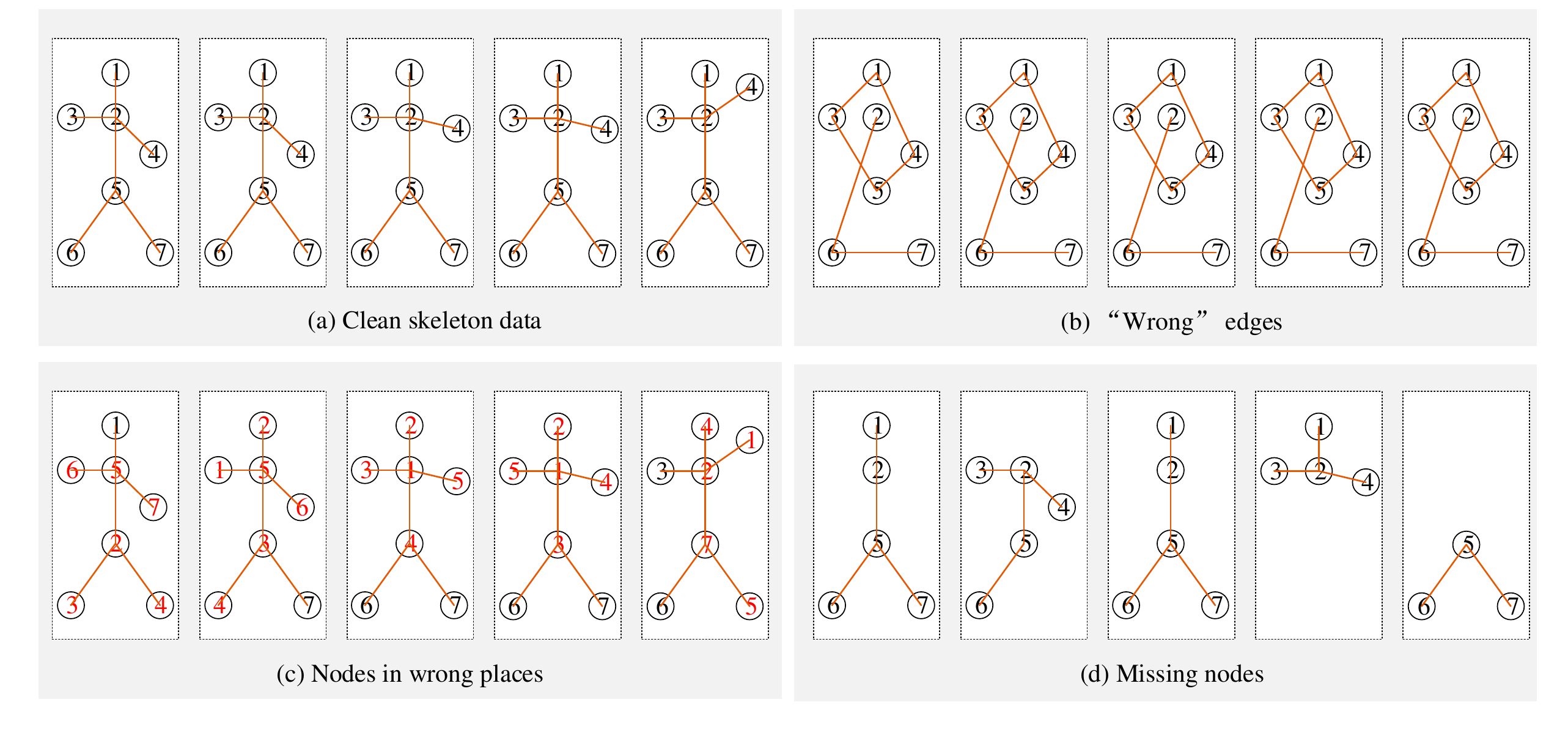}
\caption{The visualization 3 kinds of noisy skeleton data, (a) is the clean data, (b) (c) (d) are noisy data.}
\label{Fig6}
\end{figure*}
\subsection{Noise robustness analysis}
The extraction of skeleton data requires high-precision equipment and high-accuracy estimation algorithms, so lots of noise will inevitably be introduced when obtaining skeleton data. The robustness to noise is also crucial for the skeleton-based action recognition model. Recently, researchers have also proposed some models\cite{45}\cite{46} to process noisy skeleton samples. Following \cite{45}, we study the robustness of different adjacency matrices under three kinds of noise. Fig. 7 shows the examples of 3 kinds of noise.

\textbf{Noise 1: ``wrong" edges}. As Fig. 7 (b) shows, if using $A_{Skeleton}$, due to errors of humans or estimation algorithms, there will be some edges that are not connected as the human skeleton structure. Here we call these edges ``wrong" edges. We train and test some models by randomly adding different numbers of ``wrong" edges to the data. Results are shown in Fig. 8.
\begin{figure}[htb]
\centering
\includegraphics[width=3.5in]{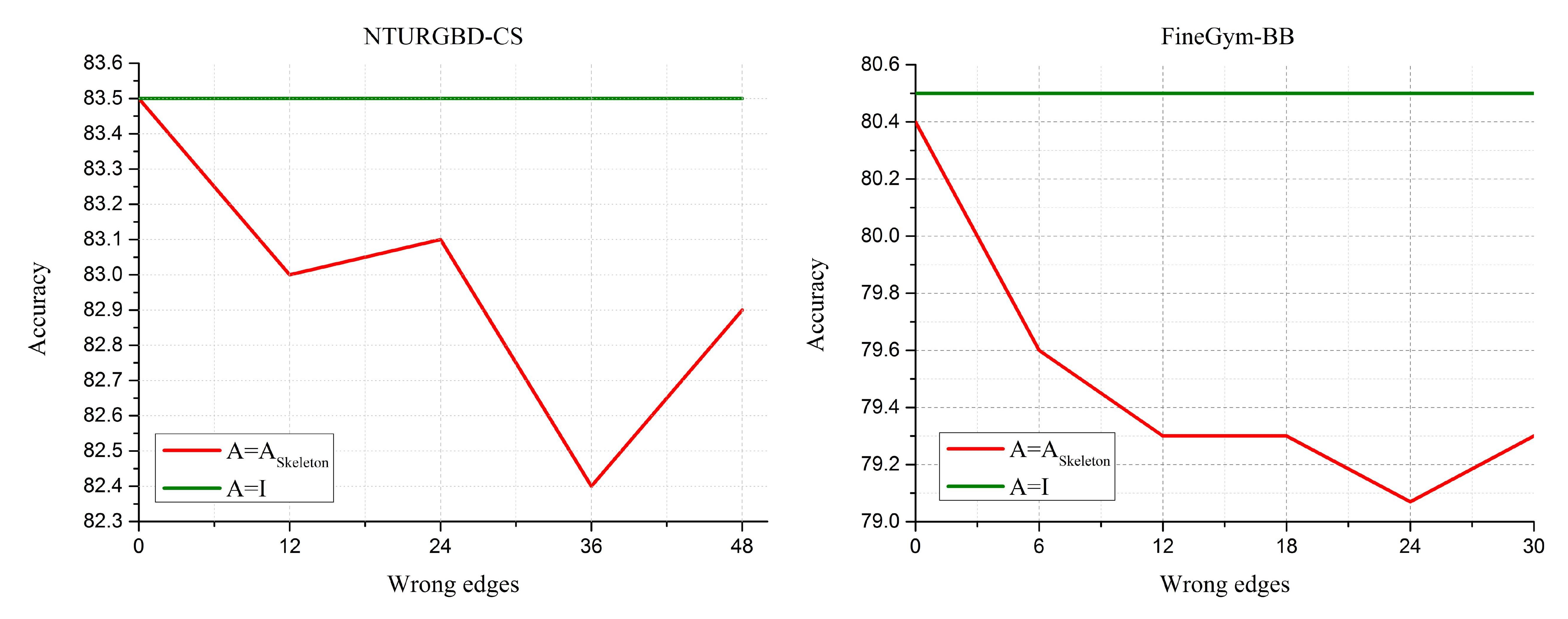}
\caption{The influence of ``wrong" edges on model accuracy(\%) under different adjacency matrices}
\label{Fig6}
\end{figure}

Fig. 8 shows that the ``wrong" edges will decrease the accuracy of models which use $A_{Skeleton}$ as the adjacency matrix. Setting $A=I$ can completely avoid the influence of this noise.

\textbf{Noise 2: Nodes in ``wrong" places}. As Fig. 7 (c) shows, due to the deformation and occlusion of the human body, the type of the skeleton node may be incorrectly estimated. We randomly shuffle the order of a different number of joints in the skeleton data and then input them into the model for training and testing. Results are shown in Fig. 9.
\begin{figure}[htb]
\centering
\includegraphics[width=3.5in]{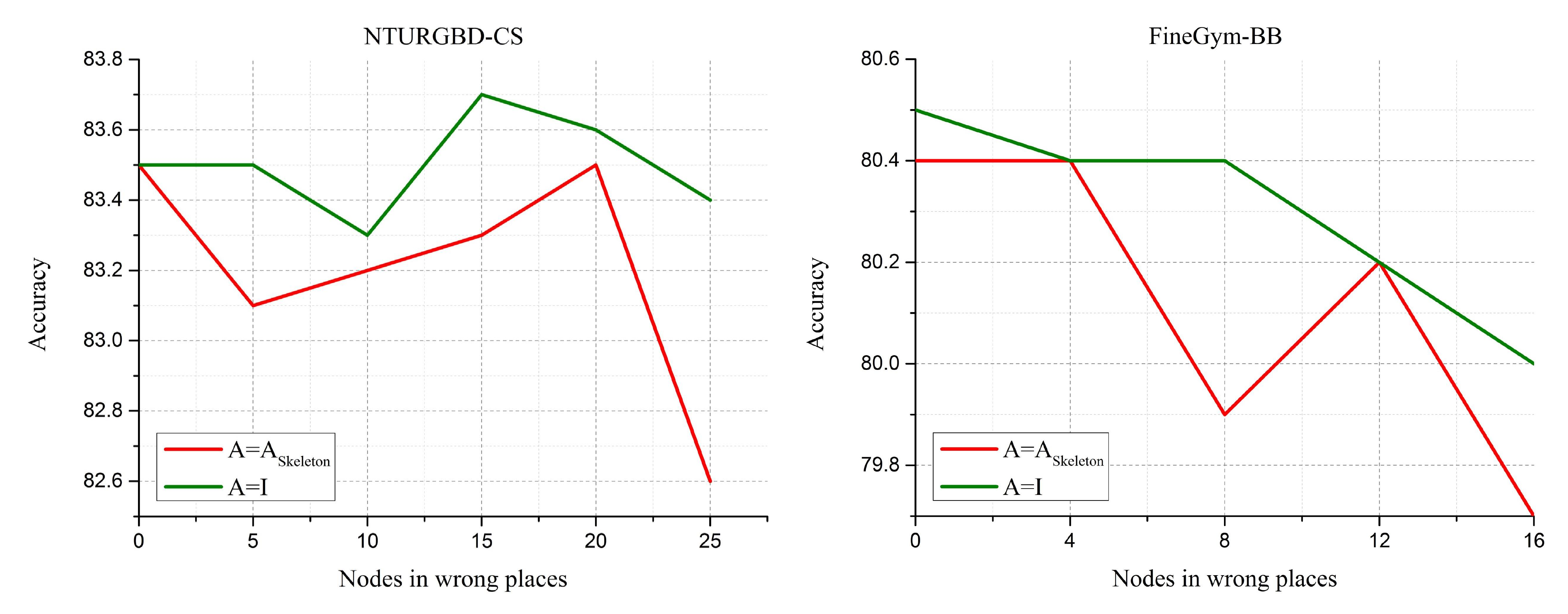}
\caption{The influence of nodes in ``wrong" places on model accuracy(\%) under different adjacency matrices}
\label{Fig6}
\end{figure}

It can be seen from Fig. 9 that shuffling the order of joints will affect the accuracy of models. But the model with $A=I$ has a smaller drop compared with $A=A_{Skeleton}$.

\textbf{Noise 3: Missing nodes}. As Fig. 7 (d) shows, some skeleton nodes will be lost due to occlusion or motion blur on the human body. We randomly drop a different number of joints in the skeleton data and then input them into the model for training and testing. The results are shown in Fig. 10.

\begin{figure}[htb]
\centering
\includegraphics[width=3.5in]{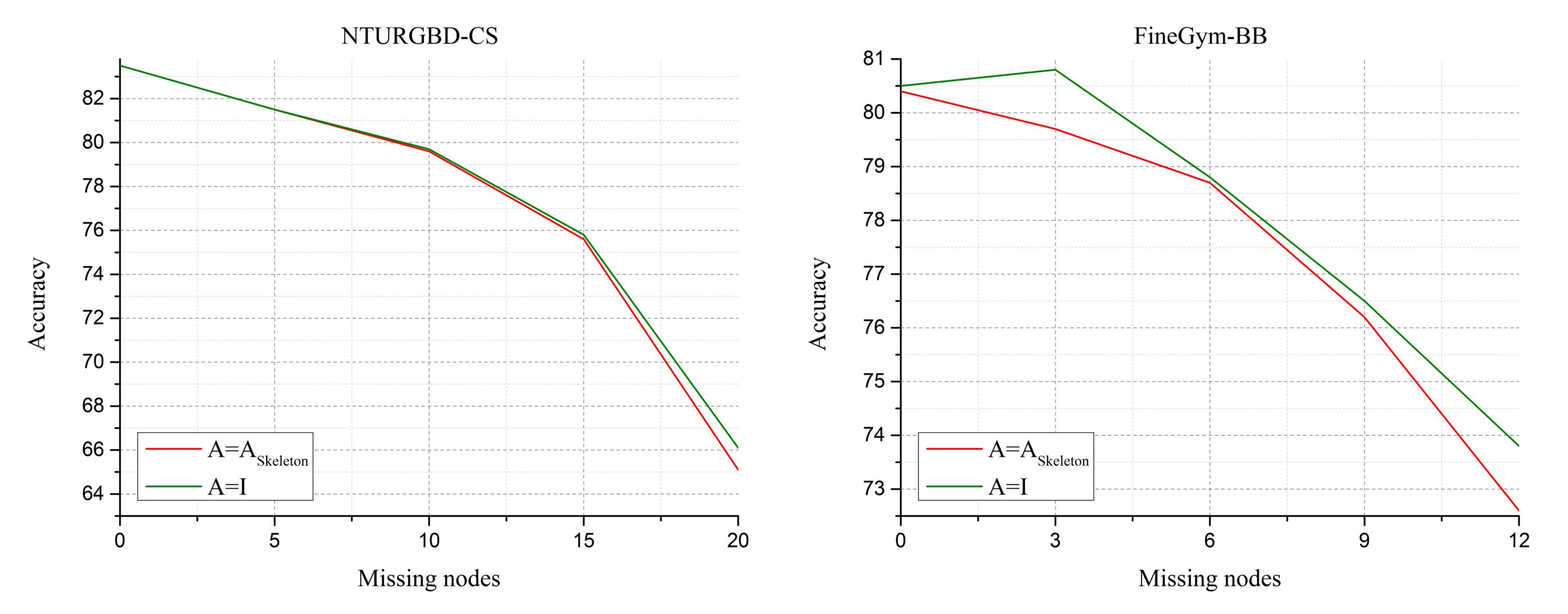}
\caption{The influence of missing nodes on model accuracy(\%) under different adjacency matrices}
\label{Fig6}
\end{figure}

The results in Fig. 10 show that the model performance decreases as the missing nodes increase. Compared $A_{Skeleton}$, setting $A=I$ is still a relatively better choice, although the advantages are not very obvious.
\begin{figure*}[htb]
\centering
\includegraphics[width=4in]{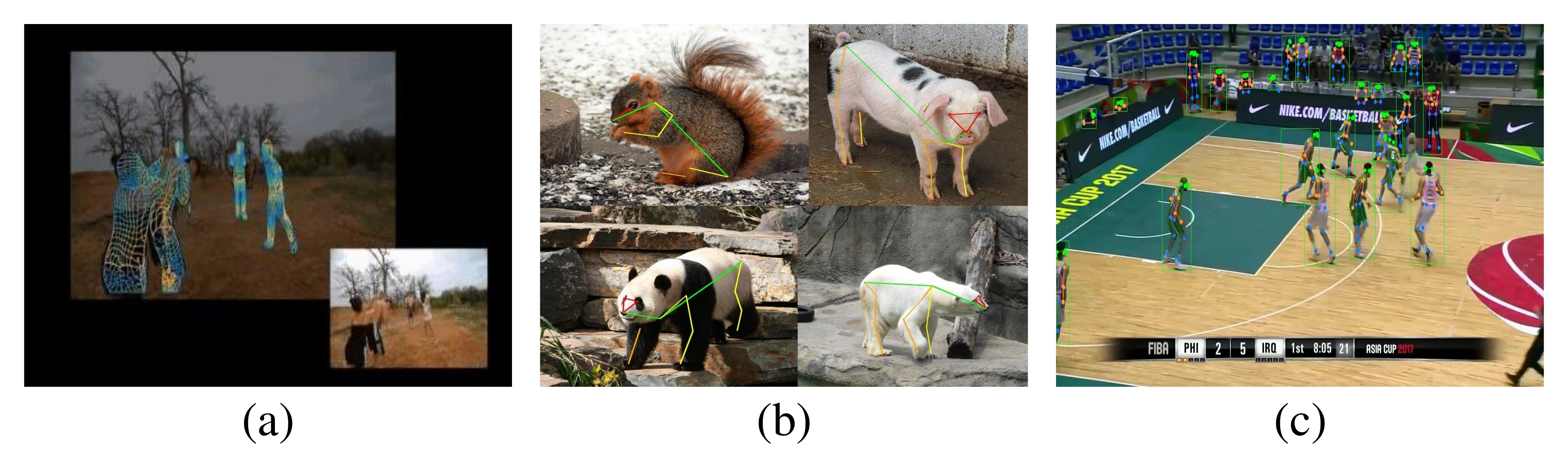}
\caption{Examples of different skeleton datasets, (a) is the dense skeleton example propose in \cite{47}, (b) shows some animal skeletons proposed in \cite{48}, (c) is a skeleton estimation result of playing basketball}
\label{Fig6}
\end{figure*}
\subsection{Transferability analysis}
In addition to the advantages of performance and noise robustness, the adjacency matrix proposed in this paper is easier to transfer to different skeleton datasets. The advantages are mainly in two aspects:

(1) \textbf{For skeleton data with different structures}: due to the diversity of estimation algorithms and equipment, many skeletons with different structures will be obtained -- for example, the two datasets Fig. 4. There are two more vivid examples in Fig. 11 (a) and (b). It is hard to define the $A_{Skeleton}$ for skeleton data like Fig. 11 (a). In the animal skeleton dataset, using $A_{Skeleton}$ is also inconvenient because of the diverse structures of different animals. Secondly, in Fig. 11 (a) and (b), the possibility of noise will significantly increased if we use $A_{Skeleton}$.
However, using the adjacency matrix $A=I+A_{res}$  does not have to worry about the above problems. For different types of skeleton data, the only information needed is the number of joints regardless of their internal structure.

(2) \textbf{For actions with multiple people}: as Fig. 11 (c) shows, for the group action, $A=I+A_{res}$ is more convenient to set. Besides, the noise in group actions is more common than actions performed by a single person, which is not suitable to use $A_{Skeleton}$.

\section{Conclusion}\label{sec13}

This paper deeply analyzes the adjacency matrix in the GCN model for skeleton-based action recognition. We propose that the most widely used $A_{Skeleton}$ is not proper in skeleton-based action recognition. Then a new adjacency matrix is proposed. Extensive experimental results and analysis demonstrate that the proposed adjacency matrix is better than $A_{Skeleton}$ in terms of performance, noise robustness, and transferability.

We think the most significant contribution of this paper is that we reveal a fact -- the natural skeleton connection does not have much effect in skeleton-based action recognition. It is a beneficial and essential conclusion for future researchers. Although we only research the GCN model, the conclusions of this paper are also applicable to other models like CNN or RNN. Because for skeleton-based action recognition, the relationships between joints are the critical factors for any method.

Regarding the future works, we think it is valuable to conduct more researches on Fig. 6. The contents in Fig. 6 are not only helpful in analyzing the adjacency matrix, but also helpful in explaining or designing the deep model and action analysis.


\bibliography{sn-bibliography}


\begin{thebibliography}{31}
\ifx \bisbn   \undefined \def \bisbn  #1{ISBN #1}\fi
\ifx \binits  \undefined \def \binits#1{#1}\fi
\ifx \bauthor  \undefined \def \bauthor#1{#1}\fi
\ifx \batitle  \undefined \def \batitle#1{#1}\fi
\ifx \bjtitle  \undefined \def \bjtitle#1{#1}\fi
\ifx \bvolume  \undefined \def \bvolume#1{\textbf{#1}}\fi
\ifx \byear  \undefined \def \byear#1{#1}\fi
\ifx \bissue  \undefined \def \bissue#1{#1}\fi
\ifx \bfpage  \undefined \def \bfpage#1{#1}\fi
\ifx \blpage  \undefined \def \blpage #1{#1}\fi
\ifx \burl  \undefined \def \burl#1{\textsf{#1}}\fi
\ifx \doiurl  \undefined \def \doiurl#1{\url{https://doi.org/#1}}\fi
\ifx \betal  \undefined \def \betal{\textit{et al.}}\fi
\ifx \binstitute  \undefined \def \binstitute#1{#1}\fi
\ifx \binstitutionaled  \undefined \def \binstitutionaled#1{#1}\fi
\ifx \bctitle  \undefined \def \bctitle#1{#1}\fi
\ifx \beditor  \undefined \def \beditor#1{#1}\fi
\ifx \bpublisher  \undefined \def \bpublisher#1{#1}\fi
\ifx \bbtitle  \undefined \def \bbtitle#1{#1}\fi
\ifx \bedition  \undefined \def \bedition#1{#1}\fi
\ifx \bseriesno  \undefined \def \bseriesno#1{#1}\fi
\ifx \blocation  \undefined \def \blocation#1{#1}\fi
\ifx \bsertitle  \undefined \def \bsertitle#1{#1}\fi
\ifx \bsnm \undefined \def \bsnm#1{#1}\fi
\ifx \bsuffix \undefined \def \bsuffix#1{#1}\fi
\ifx \bparticle \undefined \def \bparticle#1{#1}\fi
\ifx \barticle \undefined \def \barticle#1{#1}\fi
\bibcommenthead
\ifx \bconfdate \undefined \def \bconfdate #1{#1}\fi
\ifx \botherref \undefined \def \botherref #1{#1}\fi
\ifx \url \undefined \def \url#1{\textsf{#1}}\fi
\ifx \bchapter \undefined \def \bchapter#1{#1}\fi
\ifx \bbook \undefined \def \bbook#1{#1}\fi
\ifx \bcomment \undefined \def \bcomment#1{#1}\fi
\ifx \oauthor \undefined \def \oauthor#1{#1}\fi
\ifx \citeauthoryear \undefined \def \citeauthoryear#1{#1}\fi
\ifx \endbibitem  \undefined \def \endbibitem {}\fi
\ifx \bconflocation  \undefined \def \bconflocation#1{#1}\fi
\ifx \arxivurl  \undefined \def \arxivurl#1{\textsf{#1}}\fi
\csname PreBibitemsHook\endcsname

\bibitem{23}
\begin{bchapter}
\bauthor{\bsnm{Theodoridis}, \binits{T.}},
\bauthor{\bsnm{Hu}, \binits{H.}}:
\bctitle{Action classification of 3d human models using dynamic anns for mobile
  robot surveillance}.
In: \bbtitle{2007 IEEE International Conference on Robotics and Biomimetics
  (ROBIO)},
pp. \bfpage{371}--\blpage{376}
(\byear{2007}).
\bcomment{IEEE}
\end{bchapter}
\endbibitem

\bibitem{24}
\begin{bchapter}
\bauthor{\bsnm{Ren}, \binits{Z.}},
\bauthor{\bsnm{Meng}, \binits{J.}},
\bauthor{\bsnm{Yuan}, \binits{J.}},
\bauthor{\bsnm{Zhang}, \binits{Z.}}:
\bctitle{Robust hand gesture recognition with kinect sensor}.
In: \bbtitle{Proceedings of the 19th ACM International Conference on
  Multimedia},
pp. \bfpage{759}--\blpage{760}
(\byear{2011})
\end{bchapter}
\endbibitem

\bibitem{25}
\begin{bchapter}
\bauthor{\bsnm{Yang}, \binits{F.}},
\bauthor{\bsnm{Wu}, \binits{Y.}},
\bauthor{\bsnm{Sakti}, \binits{S.}},
\bauthor{\bsnm{Nakamura}, \binits{S.}}:
\bctitle{Make skeleton-based action recognition model smaller, faster and
  better}.
In: \bbtitle{Proceedings of the ACM Multimedia Asia},
pp. \bfpage{1}--\blpage{6}
(\byear{2019})
\end{bchapter}
\endbibitem

\bibitem{7}
\begin{botherref}
\oauthor{\bsnm{Liu}, \binits{H.}},
\oauthor{\bsnm{Tu}, \binits{J.}},
\oauthor{\bsnm{Liu}, \binits{M.}}:
Two-stream 3d convolutional neural network for skeleton-based action
  recognition.
arXiv preprint arXiv:1705.08106
(2017)
\end{botherref}
\endbibitem

\bibitem{4}
\begin{bchapter}
\bauthor{\bsnm{{Yong Du}}},
\bauthor{\bsnm{{Wang}}, \binits{W.}},
\bauthor{\bsnm{{Wang}}, \binits{L.}}:
\bctitle{Hierarchical recurrent neural network for skeleton based action
  recognition}.
In: \bbtitle{2015 IEEE Conference on Computer Vision and Pattern Recognition
  (CVPR)},
pp. \bfpage{1110}--\blpage{1118}
(\byear{2015}).
\doiurl{10.1109/CVPR.2015.7298714}
\end{bchapter}
\endbibitem

\bibitem{10}
\begin{bchapter}
\bauthor{\bsnm{Yan}, \binits{S.}},
\bauthor{\bsnm{Xiong}, \binits{Y.}},
\bauthor{\bsnm{Lin}, \binits{D.}}:
\bctitle{Spatial temporal graph convolutional networks for skeleton-based
  action recognition}.
In: \bbtitle{Proceedings of the AAAI Conference on Artificial Intelligence},
vol. \bseriesno{32}
(\byear{2018})
\end{bchapter}
\endbibitem

\bibitem{11}
\begin{bchapter}
\bauthor{\bsnm{Shi}, \binits{L.}},
\bauthor{\bsnm{Zhang}, \binits{Y.}},
\bauthor{\bsnm{Cheng}, \binits{J.}},
\bauthor{\bsnm{Lu}, \binits{H.}}:
\bctitle{Two-stream adaptive graph convolutional networks for skeleton-based
  action recognition}.
In: \bbtitle{Proceedings of the IEEE/CVF Conference on Computer Vision and
  Pattern Recognition},
pp. \bfpage{12026}--\blpage{12035}
(\byear{2019})
\end{bchapter}
\endbibitem

\bibitem{13}
\begin{bchapter}
\bauthor{\bsnm{Zhang}, \binits{P.}},
\bauthor{\bsnm{Lan}, \binits{C.}},
\bauthor{\bsnm{Zeng}, \binits{W.}},
\bauthor{\bsnm{Xing}, \binits{J.}},
\bauthor{\bsnm{Xue}, \binits{J.}},
\bauthor{\bsnm{Zheng}, \binits{N.}}:
\bctitle{Semantics-guided neural networks for efficient skeleton-based human
  action recognition}.
In: \bbtitle{Proceedings of the IEEE/CVF Conference on Computer Vision and
  Pattern Recognition},
pp. \bfpage{1112}--\blpage{1121}
(\byear{2020})
\end{bchapter}
\endbibitem

\bibitem{40}
\begin{barticle}
\bauthor{\bsnm{{Song}}, \binits{Y.-F.}},
\bauthor{\bsnm{{Zhang}}, \binits{Z.}},
\bauthor{\bsnm{{Shan}}, \binits{C.}},
\bauthor{\bsnm{{Wang}}, \binits{L.}}:
\batitle{Richly activated graph convolutional network for robust skeleton-based
  action recognition}.
\bjtitle{IEEE Transactions on Circuits and Systems for Video Technology}
\bvolume{31}(\bissue{5}),
\bfpage{1915}--\blpage{1925}
(\byear{2021})
\end{barticle}
\endbibitem

\bibitem{17}
\begin{bchapter}
\bauthor{\bsnm{Liu}, \binits{Z.}},
\bauthor{\bsnm{Zhang}, \binits{H.}},
\bauthor{\bsnm{Chen}, \binits{Z.}},
\bauthor{\bsnm{Wang}, \binits{Z.}},
\bauthor{\bsnm{Ouyang}, \binits{W.}}:
\bctitle{Disentangling and unifying graph convolutions for skeleton-based
  action recognition}.
In: \bbtitle{Proceedings of the IEEE/CVF Conference on Computer Vision and
  Pattern Recognition},
pp. \bfpage{143}--\blpage{152}
(\byear{2020})
\end{bchapter}
\endbibitem

\bibitem{41}
\begin{barticle}
\bauthor{\bsnm{Fang}, \binits{Z.}},
\bauthor{\bsnm{Zhang}, \binits{X.}},
\bauthor{\bsnm{Cao}, \binits{T.}},
\bauthor{\bsnm{Zheng}, \binits{Y.}},
\bauthor{\bsnm{Sun}, \binits{M.}}:
\batitle{Spatial-temporal slowfast graph convolutional network for
  skeleton-based action recognition}.
\bjtitle{IET Computer Vision}
\bvolume{16}(\bissue{3}),
\bfpage{205}--\blpage{217}
(\byear{2022})
\end{barticle}
\endbibitem

\bibitem{8}
\begin{bchapter}
\bauthor{\bsnm{Ke}, \binits{Q.}},
\bauthor{\bsnm{Bennamoun}, \binits{M.}},
\bauthor{\bsnm{An}, \binits{S.}},
\bauthor{\bsnm{Sohel}, \binits{F.}},
\bauthor{\bsnm{Boussaid}, \binits{F.}}:
\bctitle{A new representation of skeleton sequences for 3d action recognition}.
In: \bbtitle{Proceedings of the IEEE Conference on Computer Vision and Pattern
  Recognition},
pp. \bfpage{3288}--\blpage{3297}
(\byear{2017})
\end{bchapter}
\endbibitem

\bibitem{54}
\begin{botherref}
\oauthor{\bsnm{Yang}, \binits{W.}},
\oauthor{\bsnm{Zhang}, \binits{J.}},
\oauthor{\bsnm{Cai}, \binits{J.}},
\oauthor{\bsnm{Xu}, \binits{Z.}}:
Hybridnet: Integrating gcn and cnn for skeleton-based action recognition.
Applied Intelligence,
1--12
(2022)
\end{botherref}
\endbibitem

\bibitem{55}
\begin{barticle}
\bauthor{\bsnm{Sun}, \binits{Y.}},
\bauthor{\bsnm{Huang}, \binits{H.}},
\bauthor{\bsnm{Yun}, \binits{X.}},
\bauthor{\bsnm{Yang}, \binits{B.}},
\bauthor{\bsnm{Dong}, \binits{K.}}:
\batitle{Triplet attention multiple spacetime-semantic graph convolutional
  network for skeleton-based action recognition}.
\bjtitle{Applied Intelligence}
\bvolume{52}(\bissue{1}),
\bfpage{113}--\blpage{126}
(\byear{2022})
\end{barticle}
\endbibitem

\bibitem{15}
\begin{bchapter}
\bauthor{\bsnm{Cheng}, \binits{K.}},
\bauthor{\bsnm{Zhang}, \binits{Y.}},
\bauthor{\bsnm{He}, \binits{X.}},
\bauthor{\bsnm{Chen}, \binits{W.}},
\bauthor{\bsnm{Cheng}, \binits{J.}},
\bauthor{\bsnm{Lu}, \binits{H.}}:
\bctitle{Skeleton-based action recognition with shift graph convolutional
  network}.
In: \bbtitle{Proceedings of the IEEE/CVF Conference on Computer Vision and
  Pattern Recognition},
pp. \bfpage{183}--\blpage{192}
(\byear{2020})
\end{bchapter}
\endbibitem

\bibitem{16}
\begin{bchapter}
\bauthor{\bsnm{Gao}, \binits{X.}},
\bauthor{\bsnm{Hu}, \binits{W.}},
\bauthor{\bsnm{Tang}, \binits{J.}},
\bauthor{\bsnm{Liu}, \binits{J.}},
\bauthor{\bsnm{Guo}, \binits{Z.}}:
\bctitle{Optimized skeleton-based action recognition via sparsified graph
  regression}.
In: \bbtitle{Proceedings of the 27th ACM International Conference on
  Multimedia},
pp. \bfpage{601}--\blpage{610}
(\byear{2019})
\end{bchapter}
\endbibitem

\bibitem{50}
\begin{botherref}
\oauthor{\bsnm{Wu}, \binits{C.}},
\oauthor{\bsnm{Wu}, \binits{X.-J.}},
\oauthor{\bsnm{Kittler}, \binits{J.}}:
Graph2net: Perceptually-enriched graph learning for skeleton-based action
  recognition.
IEEE Transactions on Circuits and Systems for Video Technology
(2021)
\end{botherref}
\endbibitem

\bibitem{14}
\begin{bchapter}
\bauthor{\bsnm{Huang}, \binits{L.}},
\bauthor{\bsnm{Huang}, \binits{Y.}},
\bauthor{\bsnm{Ouyang}, \binits{W.}},
\bauthor{\bsnm{Wang}, \binits{L.}}:
\bctitle{Part-level graph convolutional network for skeleton-based action
  recognition}.
In: \bbtitle{Proceedings of the AAAI Conference on Artificial Intelligence},
vol. \bseriesno{34},
pp. \bfpage{11045}--\blpage{11052}
(\byear{2020})
\end{bchapter}
\endbibitem

\bibitem{49}
\begin{barticle}
\bauthor{\bsnm{Yang}, \binits{H.}},
\bauthor{\bsnm{Yan}, \binits{D.}},
\bauthor{\bsnm{Zhang}, \binits{L.}},
\bauthor{\bsnm{Sun}, \binits{Y.}},
\bauthor{\bsnm{Li}, \binits{D.}},
\bauthor{\bsnm{Maybank}, \binits{S.J.}}:
\batitle{Feedback graph convolutional network for skeleton-based action
  recognition}.
\bjtitle{IEEE Transactions on Image Processing}
\bvolume{31},
\bfpage{164}--\blpage{175}
(\byear{2022}).
\doiurl{10.1109/TIP.2021.3129117}
\end{barticle}
\endbibitem

\bibitem{26}
\begin{bchapter}
\bauthor{\bsnm{Li}, \binits{M.}},
\bauthor{\bsnm{Chen}, \binits{S.}},
\bauthor{\bsnm{Chen}, \binits{X.}},
\bauthor{\bsnm{Zhang}, \binits{Y.}},
\bauthor{\bsnm{Wang}, \binits{Y.}},
\bauthor{\bsnm{Tian}, \binits{Q.}}:
\bctitle{Actional-structural graph convolutional networks for skeleton-based
  action recognition}.
In: \bbtitle{Proceedings of the IEEE/CVF Conference on Computer Vision and
  Pattern Recognition},
pp. \bfpage{3595}--\blpage{3603}
(\byear{2019})
\end{bchapter}
\endbibitem

\bibitem{28}
\begin{bchapter}
\bauthor{\bsnm{Shahroudy}, \binits{A.}},
\bauthor{\bsnm{Liu}, \binits{J.}},
\bauthor{\bsnm{Ng}, \binits{T.-T.}},
\bauthor{\bsnm{Wang}, \binits{G.}}:
\bctitle{Ntu rgb+ d: A large scale dataset for 3d human activity analysis}.
In: \bbtitle{Proceedings of the IEEE Conference on Computer Vision and Pattern
  Recognition},
pp. \bfpage{1010}--\blpage{1019}
(\byear{2016})
\end{bchapter}
\endbibitem

\bibitem{18}
\begin{botherref}
\oauthor{\bsnm{Veli{\v{c}}kovi{\'c}}, \binits{P.}},
\oauthor{\bsnm{Cucurull}, \binits{G.}},
\oauthor{\bsnm{Casanova}, \binits{A.}},
\oauthor{\bsnm{Romero}, \binits{A.}},
\oauthor{\bsnm{Lio}, \binits{P.}},
\oauthor{\bsnm{Bengio}, \binits{Y.}}:
Graph attention networks.
arXiv preprint arXiv:1710.10903
(2017)
\end{botherref}
\endbibitem

\bibitem{42}
\begin{bchapter}
\bauthor{\bsnm{{Shao}}, \binits{D.}},
\bauthor{\bsnm{{Zhao}}, \binits{Y.}},
\bauthor{\bsnm{{Dai}}, \binits{B.}},
\bauthor{\bsnm{{Lin}}, \binits{D.}}:
\bctitle{Finegym: A hierarchical video dataset for fine-grained action
  understanding}.
In: \bbtitle{2020 IEEE/CVF Conference on Computer Vision and Pattern
  Recognition (CVPR)},
pp. \bfpage{2616}--\blpage{2625}
(\byear{2020})
\end{bchapter}
\endbibitem

\bibitem{43}
\begin{bchapter}
\bauthor{\bsnm{He}, \binits{K.}},
\bauthor{\bsnm{Gkioxari}, \binits{G.}},
\bauthor{\bsnm{Dollar}, \binits{P.}},
\bauthor{\bsnm{Girshick}, \binits{R.}}:
\bctitle{Mask r-cnn}.
In: \bbtitle{Proceedings of the IEEE International Conference on Computer
  Vision (ICCV)}
(\byear{2017})
\end{bchapter}
\endbibitem

\bibitem{44}
\begin{bchapter}
\bauthor{\bsnm{Sun}, \binits{K.}},
\bauthor{\bsnm{Xiao}, \binits{B.}},
\bauthor{\bsnm{Liu}, \binits{D.}},
\bauthor{\bsnm{Wang}, \binits{J.}}:
\bctitle{Deep high-resolution representation learning for human pose
  estimation}.
In: \bbtitle{Proceedings of the IEEE/CVF Conference on Computer Vision and
  Pattern Recognition (CVPR)}
(\byear{2019})
\end{bchapter}
\endbibitem

\bibitem{20}
\begin{bchapter}
\bauthor{\bsnm{Cui}, \binits{G.}},
\bauthor{\bsnm{Zhou}, \binits{J.}},
\bauthor{\bsnm{Yang}, \binits{C.}},
\bauthor{\bsnm{Liu}, \binits{Z.}}:
\bctitle{Adaptive graph encoder for attributed graph embedding}.
In: \bbtitle{Proceedings of the 26th ACM SIGKDD International Conference on
  Knowledge Discovery \& Data Mining},
pp. \bfpage{976}--\blpage{985}
(\byear{2020})
\end{bchapter}
\endbibitem

\bibitem{22}
\begin{bchapter}
\bauthor{\bsnm{Taubin}, \binits{G.}}:
\bctitle{A signal processing approach to fair surface design}.
In: \bbtitle{Proceedings of the 22nd Annual Conference on Computer Graphics and
  Interactive Techniques},
pp. \bfpage{351}--\blpage{358}
(\byear{1995})
\end{bchapter}
\endbibitem

\bibitem{45}
\begin{botherref}
\oauthor{\bsnm{Yoon}, \binits{Y.}},
\oauthor{\bsnm{Yu}, \binits{J.}},
\oauthor{\bsnm{Jeon}, \binits{M.}}:
Predictively encoded graph convolutional network for noise-robust
  skeleton-based action recognition.
Applied Intelligence,
1--15
(2021)
\end{botherref}
\endbibitem

\bibitem{46}
\begin{bchapter}
\bauthor{\bsnm{Song}, \binits{Y.-F.}},
\bauthor{\bsnm{Zhang}, \binits{Z.}},
\bauthor{\bsnm{Wang}, \binits{L.}}:
\bctitle{Richly activated graph convolutional network for action recognition
  with incomplete skeletons}.
In: \bbtitle{2019 IEEE International Conference on Image Processing (ICIP)},
pp. \bfpage{1}--\blpage{5}
(\byear{2019}).
\bcomment{IEEE}
\end{bchapter}
\endbibitem

\bibitem{47}
\begin{bchapter}
\bauthor{\bsnm{G{\"u}ler}, \binits{R.A.}},
\bauthor{\bsnm{Neverova}, \binits{N.}},
\bauthor{\bsnm{Kokkinos}, \binits{I.}}:
\bctitle{Densepose: Dense human pose estimation in the wild}.
In: \bbtitle{Proceedings of the IEEE Conference on Computer Vision and Pattern
  Recognition},
pp. \bfpage{7297}--\blpage{7306}
(\byear{2018})
\end{bchapter}
\endbibitem

\bibitem{48}
\begin{botherref}
\oauthor{\bsnm{Yu}, \binits{H.}},
\oauthor{\bsnm{Xu}, \binits{Y.}},
\oauthor{\bsnm{Zhang}, \binits{J.}},
\oauthor{\bsnm{Zhao}, \binits{W.}},
\oauthor{\bsnm{Guan}, \binits{Z.}},
\oauthor{\bsnm{Tao}, \binits{D.}}:
Ap-10k: A benchmark for animal pose estimation in the wild.
arXiv preprint arXiv:2108.12617
(2021)
\end{botherref}
\endbibitem

\end{thebibliography}


\end{document}